\documentclass[letterpaper]{article} 
\usepackage{aaai2026}  
\usepackage{times}  
\usepackage{helvet}  
\usepackage{courier}  
\usepackage[hyphens]{url}  
\usepackage{graphicx} 
\usepackage{natbib}  
\usepackage{caption} 
\usepackage{algorithm}
\usepackage{algorithmic}
\usepackage{amsmath}
\usepackage{amssymb}
\usepackage{mathtools}
\usepackage{amsthm}
\usepackage{adjustbox}
\usepackage{booktabs, multirow}
\usepackage{colortbl}

\newcommand{\FacialReactionGenerator}{FRG}
\newcommand{\personalisedbehaviourPatternLearning}{PBPL}
\newcommand{\PersonalWeightGenerator}{PWHG}
\newcommand{\MatrixGraphNeuralNetwork}{2D-GNN} 

\newcommand{\personalisedCognitionSimulation}{PCS}
\newcommand{\RealPersonalityRecognition}{RPR}

\usepackage{newfloat}
\usepackage{listings}
\DeclareCaptionStyle{ruled}{labelfont=normalfont,labelsep=colon,strut=off} 
\floatstyle{ruled}
\newfloat{listing}{tb}{lst}{}
\floatname{listing}{Listing}
\title{Learning Personalised Human Internal Cognition from External Expressive Behaviours for Real Personality Recognition}
\author{
    Xiangyu Kong\textsuperscript{\rm 1},
    Hengde Zhu\textsuperscript{\rm 2},
    Haoqin Sun\textsuperscript{\rm 3},
    Zhihao Guo\textsuperscript{\rm 4},
    Jiayan Gu\textsuperscript{\rm 2},
    Xinyi Ni\textsuperscript{\rm 1},
    Wei Zhang\textsuperscript{\rm 5},\\
    Shizhe Liu\textsuperscript{\rm 6},
    Siyang Song\textsuperscript{\rm 1}\thanks{Corresponding author.},
}

\affiliations{
    \textsuperscript{\rm 1} Department of Computer Science, University of Exeter\\
    \textsuperscript{\rm 2} School of Computing and Mathematical Sciences, University of Leicester\\
    \textsuperscript{\rm 3} College of Computer Science, Nankai University\\
    \textsuperscript{\rm 4} The Manchester Metropolitan University\\
    \textsuperscript{\rm 5} School of Software Technology, Zhejiang University\\
    \textsuperscript{\rm 6} Department of Computer Science, University of Oxford\\

}

\usepackage{bibentry}

\begin{document}

\maketitle

\begin{abstract}
Automatic real personality recognition (RPR) aims to evaluate human real personality traits from their expressive behaviours. However, most existing solutions generally act as external observers to infer observers' personality impressions based on target individuals' expressive behaviours, which significantly deviate from their real personalities and consistently lead to inferior recognition performance. Inspired by the association between real personality and human internal cognition underlying the generation of expressive behaviours, we propose a novel RPR approach that efficiently simulates personalised internal cognition from easy-accessible external short audio-visual behaviours expressed by the target individual. The simulated personalised cognition, represented as a set of network weights that enforce the personalised network to reproduce the individual-specific facial reactions, is further encoded as a novel graph containing two-dimensional node and edge feature matrices, with a novel 2D Graph Neural Network (2D-GNN) proposed for inferring real personality traits from it. To simulate real personality-related cognition, an end-to-end strategy is designed to jointly train our cognition simulation, 2D graph construction, and personality recognition modules. Experiments show our approach’s effectiveness in capturing real personality traits with superior computational efficiency. Our code is provided in Supplementary Material.
\end{abstract}

\begin{figure}[h]
  \centering
  \includegraphics[width=\linewidth]{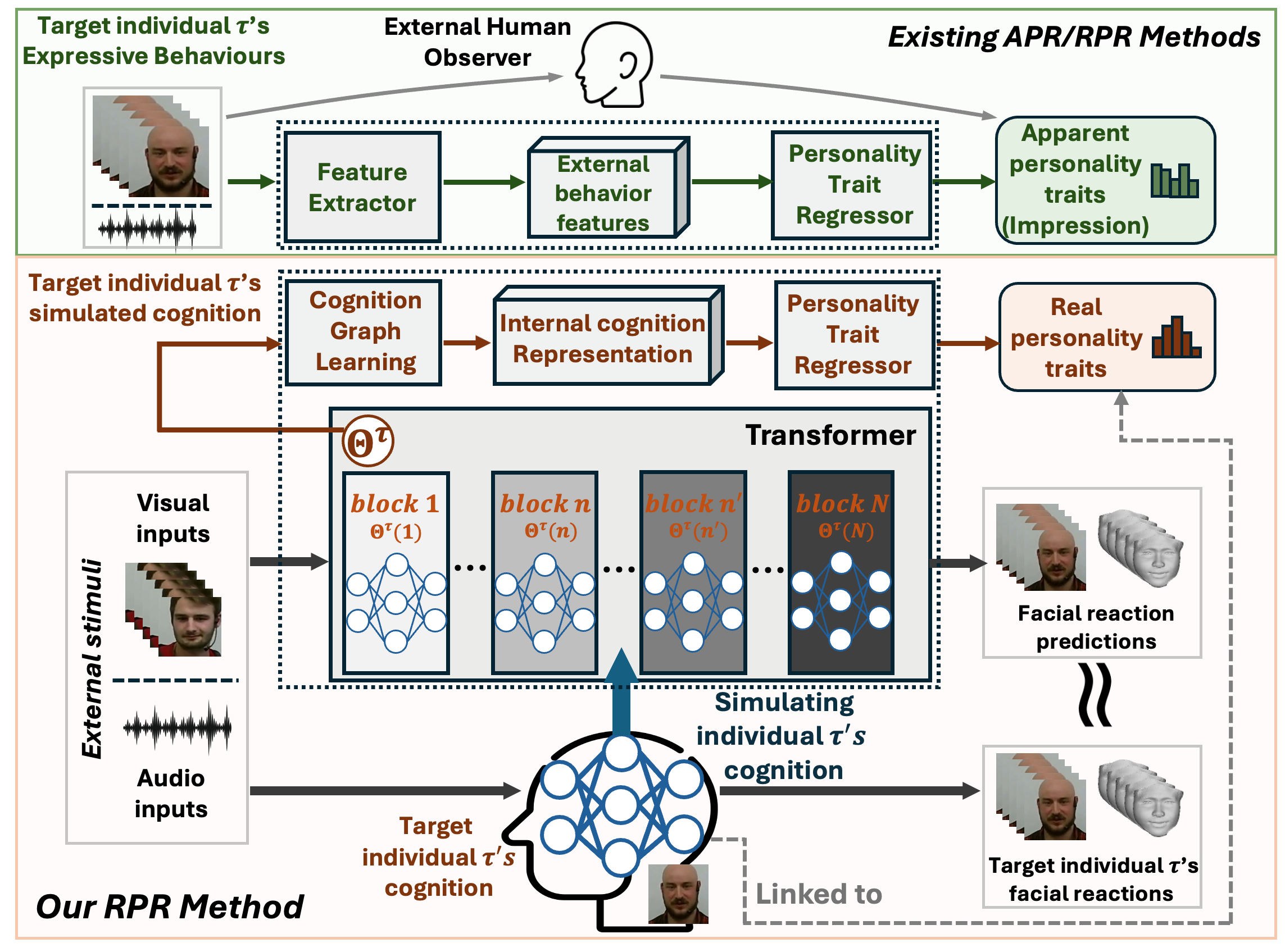}
  \caption{\textbf{Upper:} Existing methods play the role of an \textbf{external observer} to directly infer \textbf{personality impressions from the individual's \textit{external behaviours}}. \textbf{Lower:} Our novel method simulates personalised internal cognition $\Theta^\tau$ of individual $\tau$ by enforcing a personalised network $\text{FRG}_{\Theta^\tau}$ to reproduce $\tau$'s facial reactions in response to audio-visual inputs, and then infers \textbf{real personality from the simulated \textit{internal cognition} $\Theta^\tau$}.}
  \label{fig:compare}
\end{figure}

\section{Introduction}
\label{sec:intro}

Human personality represents distinctive patterns of thoughts, emotions and behaviours defining individuals \cite{corr2020cambridge}, shaped by intricate cognitive processes \cite{vandenbos2007apa}. While accurately understanding human real personality is crucial for various real-world applications such as personalizing human-computer interaction (HCI) systems \cite{norman2017cyberpsychology}, job recruitment \cite{cole2009recruiters} and mental illness diagnosis \cite{jaiswal2019automatic}, traditional self-reported questionnaires or projective tests-based personality recognition are frequently limited by subjective self-assessments and psychologists' inconsistent interpretations \cite{tuber2012understanding,mcdonald2008measuring}.

Since personality can be partially reflected by human expressive behaviours (e.g., facial expressions and speeches \cite{naumann2009personality}), previous studies have developed automatic, objective and repeatable models according to individuals' non-verbal facial or/and audio behaviours \cite{li2020cr,9373959,Curto_2021_ICCV,8897617,dodd2023framework,salam2022learning,SUMAN2022107715}. These approaches can be categorized into two types \cite{vinciarelli2014survey}: (1) apparent personality recognition (APR) predicting human observers’ impressions on the target individual; and (2) real personality recognition (RPR) predicting the individual's real personality traits related to their internal cognition \cite{ajzen2012attitudes,daly1987personality}. Despite different goals, these APR \cite{ryumina2024gated,9373959,li2020cr} and RPR \cite{salam2022learning,palmero2021context}) approaches generally infer personality directly from the target individual's observable audio-facial behaviours, which play the role of external observers to provide \textbf{\textit{impressions that are largely different from real personality traits}}, i.e., they are theoretically designed to predict apparent personality traits. This is evidenced by the poor performances of existing APR and RPR models in real personality recognition \cite{10428080}.

Inspired by the insights that human internal cognition \cite{farmer2019cognition}, which is intrinsically linked to real personality, shapes their external speech and facial behaviours \cite{corr2020cambridge,9993801,10.1145/3474085.3475460}, this paper hypothesises that existing deep neural networks (DNNs), designed with the inspiration from biological neural systems within human brain \cite{yegnanarayana2009artificial,krogh2008artificial}, can \textit{\textbf{partially approximate human cognition}} by computationally task-performing based on human-like input-output mappings (inspired by the 'Weak AI' hypothesis \cite{flowers2019strong}).
In this sense, we innovatively propose to learn a set of personalised weights that enforces a pre-defined network to reproduce personalised facial reactions (output) expressed by the target individual when perceiving the same external stimulus (input). Consequently, the learned personalised weights would moderately simulate the individual's internal cognition, which is further leveraged to infer real personality, i.e., our approach is theoretically designed to predict real personality (illustrated in Fig. \ref{fig:compare}). Unlike prior time-consuming Neural Architecture Search-based cognition simulation strategies \cite{9993801,10.1145/3474085.3475460} that require personalised network exploring/training on a long dyadic interaction clip pair (i.e., not recordable by the target individual alone) at the inference stage, our approach only performs efficient forward inference of the personalised weights based on an easy-accessible and short audio-facial clip of the individual without training.

To facilitate inferring real personality from human cognition, our approach further encodes the simulated personalised cognition to a processable cognition graph representation, allowing crucial structural information within network weights to be retained, which is finally fed to our proposed 2D Graph Neural Network (2D-GNN) for RPR. The main contributions and novelties (more details are in Supplementary Material) are summarised as follows:
\begin{itemize}
    \item We propose a novel RPR framework that end-to-end encodes the target individual’s external audio-facial behaviours into a 2D graph representation representing this individual's internal cognition, where an efficient yet effective personalised cognition simulation strategy and a 2D Graph Neural Network are proposed.

    \item We propose a novel end-to-end training strategy that jointly optimises the cognition simulation and real personality prediction modules, enabling our framework to effectively simulate real personality-related personalised cognition cues from easy-accessible human external audio-visual behaviours.

    \item Experiments not only demonstrate that our framework is the new state-of-the-art in recognising real personality traits with significant advantages over methods of traditional framework and 2,000 times inference efficiency compared to the previous cognition simulation strategy.
\end{itemize}

\section{Related Work}
\label{sec:related}

\textbf{Automatic personality recognition:} Audio-visual automatic personality recognition methods, encompassing both apparent personality recognition \cite{gurpinar2016combining,li2020cr,Curto_2021_ICCV,SUMAN2022107715,dodd2023framework} and real personality recognition \cite{salam2022learning,palmero2021context}, have been widely developed based on the pipeline which directly extracts expressive behavioural features from human face \cite{8897617,9373959,ventura2017interpreting}, body \cite{romeo2021predicting,sonlu2024towards}, speech \cite{liu2020speech,leekha2024vyaktitvanirdharan} or their combinations \cite{salam2022learning,li2020cr,ghassemi2023unsupervised}. A typical example is the CR-Net \cite{li2020cr} which leverages a ResNet-34 \cite{He_2016_CVPR} to directly extract personality features from human visual, audio and text behaviours. Alternatively, a recent approach \cite{9993801,10.1145/3474085.3475460} attempts to simulate human internal cognition via their expressive behaviours (i.e., exploring a personalised network with a unique architecture to represent each individual's cognition), and then infers real personality traits based on the simulated personalised cognition. Consequently, this approach is hindered by extremely high computational costs of personalised network exploration and dependency on long-duration dyadic interaction data at the inference stage.

\begin{figure*}[h]
  \centering
  \includegraphics[width=\linewidth]{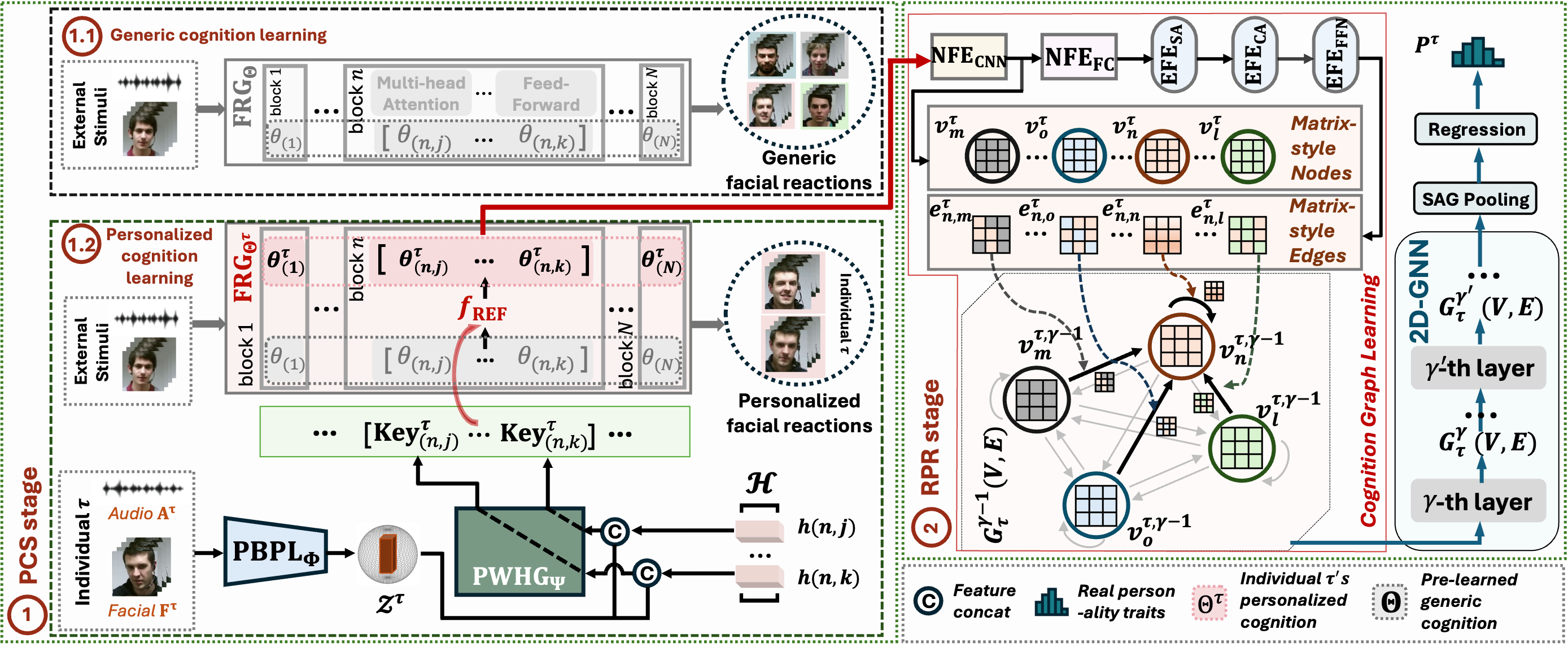}
  \caption{The pipeline of our method, which consists of two main stages: (1) Personalised Human Cognition Simulation (PCS) including (1.1) generic human cognition learning and (1.2) personalised cognition learning; and (2) novel cognition graph representation learning and 2D Graph Neural Network (2D-GNN)-based real personality recognition (detailed in \textbf{Methodology overview} section).} 
  \label{fig:pipeline}
  \vskip -0.1in
\end{figure*}

\textbf{Human cognition and real personality:} Human cognition refers to human mental processes such as understanding and planning. Various psychological studies \cite{corr2020cambridge,soubelet2011personality} claimed that human cognition is strongly associated with their real personality traits. Specifically, Extraversion and Neuroticism traits are typically linked with specific biases in human cognitive functions, contributing to either extroverted social behaviours or increased threat awareness for high neurotic individuals \cite{boyle2008personality}. Meanwhile, some neuroscience studies \cite{adelstein2011personality,deyoung2010testing} suggest that distinct brain cognitive patterns are associated with specific personality traits, emphasising how real personality modulates both emotional and cognitive responses to various external stimuli.


\section{Methodology}
\label{sec:method}

\textbf{Hypothesis:} Previous studies claimed that deep neural networks (DNNs) can link certain cognitive functions (e.g., visual object recognition) to human brain \cite{kriegeskorte2018cognitive,spoerer2017recurrent}, while some DNNs can approximate human cognition to perform realistic reasoning from languages \cite{binz2025foundation} or multi-modal sensory inputs \cite{ye2018survey,panella2021deep,9993801}. Given that Transformer mirrors hierarchical information processing functions corresponding to various human brain regions \cite{Yang_2024_CVPR}, we hypothesise that our model composed of stacked transformer blocks is capable of approximately simulating human cognition.

\textbf{Methodology overview:} As depicted in Fig. \ref{fig:pipeline}, our approach is structured in two stages to recognise the target individual $\tau$'s real personality traits from an external expressive behaviour expressed by $\tau$. The \textbf{Personalised Cognition Simulation (PCS)} stage first learns a generic human cognition $\Theta$ (a set of network weights) defining a transformer-based generic facial reaction generator $\text{\FacialReactionGenerator}_{\Theta}$ capable of generating appropriate facial reactions in response to each input (perceived) stimuli, which are plausible to be expressed by humans with varying real personalities (Fig. \ref{fig:pipeline} (1.1)).
Then, an encoder-decoder (PBPL and PWHG in Fig. \ref{fig:pipeline}) is developed to generate a personalised key $\text{Key}^{\tau}$ (e.g., a set of weight offsets that have the same form as $\Theta$) from the individual $\tau$'s short audio and facial behaviour $\{\mathrm{A}^{\tau}, \mathrm{F}^{\tau}\}$, which refines $\Theta$ as $\Theta^{\tau}$ (a set of network weights) representing the individual $\tau$'s personalised cognition through the operation $f_{\text{REF}}$ (Fig. \ref{fig:pipeline} (1.2)) as:
\begin{equation} 
   \Theta^{\tau} = f_{\text{REF}}(\text{Key}^{\tau}, \Theta)
\label{eq:key-generation}
\end{equation}
where different REF operations are evaluated in Table \ref{tab:ablation-indexing}. Here, $\Theta^{\tau}$ is learned to re-define the generic generator $\text{\FacialReactionGenerator}_{\Theta}$ as a personalised facial reaction generator $\text{\FacialReactionGenerator}_{\Theta^{\tau}}$, enforcing it to reproduce only $\tau$'s personalised facial reactions. Since the pre-obtained generic cognition $\Theta$ can always be re-used, the $\Theta^{\tau}$ of any individual $\tau$ can be efficiently achieved from a short audio-visual behaviour expressed by $\tau$.

Subsequently, the \textbf{Real Personality Recognition (\RealPersonalityRecognition)} stage starts with a \textbf{Cognition Graph Learning (CGL)} module to innovatively encode $\Theta^{\tau}$ (a set of personalised weights within the generator $\text{\FacialReactionGenerator}_{\Theta^{\tau}}$) that \textit{cannot be directly processed by standard deep learning (DL) models} as a \textit{processable personalised cognition graph $G^{\tau}(V,E)$} as: 
\begin{equation}
    G^{\tau}(V,E) = \textbf{CGL}(\Theta^{\tau})
\label{eq:graph-representation}
\end{equation}
where $G^{\tau}(V,E)$ is made up of a set of two-dimensional matrix-valued nodes $V$ and edges $E$ that retain crucial structural cues of $\text{\FacialReactionGenerator}_{\Theta^{\tau}}$'s weights $\Theta^{\tau}$ as well as their relationships. Finally, a novel 2D-GNN (\MatrixGraphNeuralNetwork) that can process nodes and edges characterized by 2D feature matrices within $G^{\tau}(V,E)$ is proposed to predict real personality traits of $\tau$ from the cognition graph $G^{\tau}(V,E)$ (Fig. \ref{fig:pipeline} (2)) as:
\begin{equation}
    \mathcal{P}^{\tau} = \textbf{\MatrixGraphNeuralNetwork}_{\varphi}(G^{\tau}(V,E))
\end{equation}

In this paper, we propose a novel joint training strategy to optimise the entire framework including \FacialReactionGenerator$_{\Theta^\tau}$, \personalisedbehaviourPatternLearning, \PersonalWeightGenerator, CGL, and \MatrixGraphNeuralNetwork\ in an end-to-end manner, allowing the well-trained \personalisedbehaviourPatternLearning\ and \PersonalWeightGenerator\ to specifically simulate real personality-related personalised cognition from the target individual's external behaviours.

\subsection{Personalised Cognition Simulation}
Since the amount of behavioural data exhibited by a single individual may be limited while individuals of different personalities sharing certain common cognitive processes, the \textbf{\personalisedCognitionSimulation\ stage} starts with pre-learning of generic human cognition $\Theta$ of varying individuals by training FRG to reproduce facial reactions expressed by different individuals in response to input audio-facial behaviours. This allows an efficient inference stage to be achieved by only learning a set of personalised $\text{Key}^{\tau}$ from relatively small amounts of behaviours expressed by the target individual $\tau$, \textit{e.g., a short audio-facial clip $\{\mathrm{A}^{\tau}, \mathrm{F}^{\tau}\}$ that can be easily recorded in non-invasive manners}, where $\text{Key}^{\tau}$ is expected to refine the pre-obtained $\Theta$ as the target personalised cognition $\Theta^\tau$,
Specifically, we first train the generic facial reaction generator $\text{\FacialReactionGenerator}_{\Theta}$  (i.e., simulating generic cognition) using dyadic audio-facial human-human interaction clips expressed by various pairs of individuals, with the objective to output appropriate facial reactions $\mathcal{R} = \{R_1, R_2, \cdots, R_L \}$ expressed by individuals of different personalities in response to each input human behaviour $\{ \mathrm{A}^{s}, \mathrm{F}^{s} \}$ as:
\begin{equation}
     \mathcal{R} = \text{\FacialReactionGenerator}_{\Theta}(\mathrm{A}^{s}, \mathrm{F}^{s})
\end{equation}
where $\Theta$ is learned as a set of network weights. To validate our hypothesis and ensure reproducibility, the $\text{\FacialReactionGenerator}_{\Theta}$ simply stacks $N$ transformer blocks with each consisting of a self-attention, a cross-attention, and a feed-forward layer containing two FC layers, which includes $K$ weights (i.e., 2D matrices for all linear layers).


Then, the \textbf{Personalised Behaviour Pattern Learning (\personalisedbehaviourPatternLearning)} module learns a latent representation $\mathcal{Z}^{\tau}$ to describe the individual $\tau$'s \textit{personalised external behaviour patterns} $\mathcal{Z}^{\tau}$ (a feature vector) from individual's \textit{short audio-facial behaviour clip} $\{\mathrm{A}^{\tau}, \mathrm{F}^{\tau}\}$ as:
\begin{equation}
      \mathcal{Z}^{\tau} = \textbf{\personalisedbehaviourPatternLearning}(\mathrm{A}^{\tau}, \mathrm{F}^{\tau})
\label{eq:PSL}
\end{equation}
where \personalisedbehaviourPatternLearning\ is a transformer-based encoder. Subsequently, a \textbf{Personalised Weight HyperGenerator (\PersonalWeightGenerator)} consisting of FC layers further projects $\mathcal{Z}^{\tau}$ to the personalised key $\text{Key}^{\tau}$ (e.g., a set of weight offsets) to refine the generic cognition $\Theta$ as the personalised cognition $\Theta^{\tau}$ as Eq. \ref{eq:key-generation}. To efficiently learn $K \times N$ matrices in $\Theta$ ($K$ matrices for parameterizing all linear layers in each of the $N$ transformer blocks), our \PersonalWeightGenerator\ module samples $N \times K$ learnable chunk embeddings $\mathcal{H} = \{ h_{n,k} | n=1, \cdots, N; k = 1, \cdots, K \} $ from the standard Gaussian distribution, and then concatenates each $h_{n, k}$ with the learned $\mathcal{Z}^{\tau}$. This results in $K \times N$ different concatenated embeddings, which are then projected to $K \times N$ matrices as $\text{Key}^{\tau}$ via a shared \PersonalWeightGenerator\ module as:
\begin{equation}
\begin{split}
    \text{Key}^{\tau} &= \{[\text{\PersonalWeightGenerator}(\mathcal{Z}^{\tau}, h_{n,k})]\}_{{n=1:N},{k=1:K}}\\
    &= \{[\text{Key}^{\tau}_{n,k}]\}_{{n=1:N},{k=1:K}}
\end{split}
\label{eq:PWG}
\end{equation}
This way, the \PersonalWeightGenerator\ only includes one shared projector for generating all weight matrices, resulting in approximately $K \times N$-fold reduction in the number of learnable weights compared to directly using $K \times N$ independent projectors.

At the inference stage, since the pre-learned $\Theta$ can always be re-used without training (i.e., the competitor \cite{9993801} requires the time-consuming personalised network exploration for every individual during inference), our approach is efficient as it only involves forwardly propagating the given audio-facial clip $\mathrm{A}^{\tau}, \mathrm{F}^{\tau}$ to refine $\Theta$ as $\Theta^{\tau}$. 

\subsection{Cognition-based real personality recognition}
Since the simulated personalised cognition $\Theta^{\tau}$ is a set of unorganised weights that are difficult to be directly processed by standard DL models for RPR, the \RealPersonalityRecognition\ stage encodes the $\Theta^{\tau}$ as an organised and processable graph $G^{\tau}$ containing node and edge feature matrices. Then, a novel 2D Graph Neural Network (\MatrixGraphNeuralNetwork) is proposed to infer real personality traits from $G^{\tau}$.

\textbf{Cognition graph representation (nodes and edges) learning:} Given $\Theta^\tau$, the RPR stage applies a \textbf{Cognition Graph Learning (CGL)} module consisting of a node feature encoding (NFE) block and an edge feature encoding (EFE) block to encode the weight matrices parameterizing all linear layers in $N$ transformer blocks of $\text{\FacialReactionGenerator}_{\Theta^{\tau}}$ to a personalised cognition graph $G^{\tau}(V,E)$, where $V = \{v_1, \cdots, v_N\}$ denotes $N$ nodes and $E = \{e_{n,m}\ |\ n, m = 1, \dots, N\}$ denotes a set of directed edges connecting all node pairs.
Specifically, the \textbf{NFE} block encodes the $K$ equal-size weights $\Theta^{\tau}_{n} = [\Theta^{\tau}_{n,1}, \cdots, \Theta^{\tau}_{n,K}]$ of the $n$-th transformer block into a single node feature $v^{\tau}_n \in \mathbb{R}^{d_v \times d_v}$ as:
\begin{align}
     v^{\tau}_n = \text{NFE}_{\text{FC}}({\text{NFE}}_{\text{CNN}}(\Theta^{\tau}_{n}))
\end{align}
It first processes every $\Theta^{\tau}_{n} \in \mathbb{R}^{K \times d \times d}$ through a CNN extractor ${\text{NFE}}_{\text{CNN}}$ to obtain the feature representation $\eta^{\tau}_n$, which is then processed by a FC layer ${\text{NFE}}_{\text{FC}}$ to yield the matrix-valued node $v^{\tau}_n$ (i.e., weighted sum of $K$ layer-level weights). This way, the final obtained $G^{\tau}$ preserves crucial structural information within weights, avoiding information loss/distortion when reducing them to vectors (evaluated in Table \ref{tab:ablation-main}). To comprehensively explore real personality-related relationships among the simulated cognitive processes (i.e., transformer blocks' weights represented by graph node representations), our \textbf{EFE} block learns a pair of edge feature matrices $e^{\tau}_{n,m}$ and $e^{\tau}_{m,n}$ between every pair of nodes $v^{\tau}_n$ and $v^{\tau}_m$. This strategy is proposed not only because it can explore all potential personality-related relationship cues among transformer blocks (i.e., the exact relationship among nodes remains unclear), but also the widely-used single-value edge features fail to comprehensively model complex relationships between a pair of node features unless they are linearly dependent. Specifically, our EFE applies a self-attention ($\text{EFE}_{\text{SA}}$) and a cross-attention ($\text{EFE}_{\text{CA}}$) to emphasise real personality-related relationships shared by every pair of node features, followed by a feed-forward network ($\text{EFE}_{\text{FFN}}$) for further refinement. These result in a pair of edge features $e^{\tau}_{n,m} \in \mathbb{R}^{d_v \times d_v}$ and $e^{\tau}_{m,n} \in \mathbb{R}^{d_v \times d_v}$ as:
\begin{align}
     e^{\tau}_{n,m} &= \text{EFE}(\eta^{\tau}_{n}, \eta^{\tau}_{m}) \nonumber \\ 
     &= \text{EFE}_{\text{FFN}}(\text{EFE}_{\text{CA}}(\text{EFE}_{\text{SA}}(\eta^{\tau}_{n}), \text{EFE}_{\text{SA}}(\eta^{\tau}_{m})))
\end{align}
More model and implementation details of the proposed NFE and EFE are provided in the Supplementary Material.

\textbf{2D Graph Neural Network (2D-GNN)}: While existing standard GNNs are limited to processing graphs with vector-based node and edge features, our novel \MatrixGraphNeuralNetwork\ directly processes node feature matrices $V$ and edge feature matrices $E$ within the learned cognition graph $G^{\tau}(V,E)$ to infer $\tau$'s real personality traits $\mathcal{P}_{\text{traits}}^{\tau}$. Specifically, the $\gamma$-th \MatrixGraphNeuralNetwork\ layer updates its input node feature $v_n^{\tau,\gamma-1} \in \mathbb{R}^{d_v \times d_v}$ within the graph $G^{\tau}(V,E)^{\gamma-1}$ as an node feature $v_n^{\tau,\gamma} \in \mathbb{R}^{d_v \times d_v}$ of the same size as:
\begin{align}
    v_n^{\tau,\gamma} &= \text{ReLU} \Big( \frac{1}{d_n} \sum_{m \in \mathcal{N}(n)} A^{\tau}_{n, m} \cdot \hat{e}^{\tau}_{n,m} \odot \hat{v}_n^{\tau,\gamma-1} \Big), \nonumber \\
    \hat{e}^{\tau}_{n,m} &= f^{(\gamma-1)}_e(e^{\tau}_{n,m}), \ \hat{v}_n^{\tau,\gamma-1} = f^{(\gamma-1)}_v(v_n^{\tau,\gamma-1})
\end{align}
where $f^{(\gamma-1)}_e$ and $f^{(\gamma-1)}_v$ are linear projections; $d_n$ is the degree of node $v_n^{\tau, \gamma-1}$ for normalization; $\odot$ represents Hadamard product, $A^{\tau}_{n, m} \in \{0, 1\}$ is binary adjacency indicator and $\mathcal{N}(n)$ denotes the $v_n^{\tau, \gamma-1}$'s neighboring nodes. This way, each node $v_n^{\tau,\gamma}$ is obtained by considering both its previous states $v_n^{\tau,\gamma-1}$ and the messages passed from its neighbouring nodes $\mathcal{N}(n)$ via the corresponding edge feature matrices. Finally, a self-attention graph pooling \cite{pmlr-v97-lee19c} integrates all updated node features, based on which three FC layers equipped with ReLU and dropout are finally employed to simultaneously infer five real personality traits.

\subsection{End-to-end joint training strategy\label{training:end-to-end}}

\begin{figure}[t]
  \centering
  \hspace{-0.04\columnwidth}
  \includegraphics[width=1.03\columnwidth]{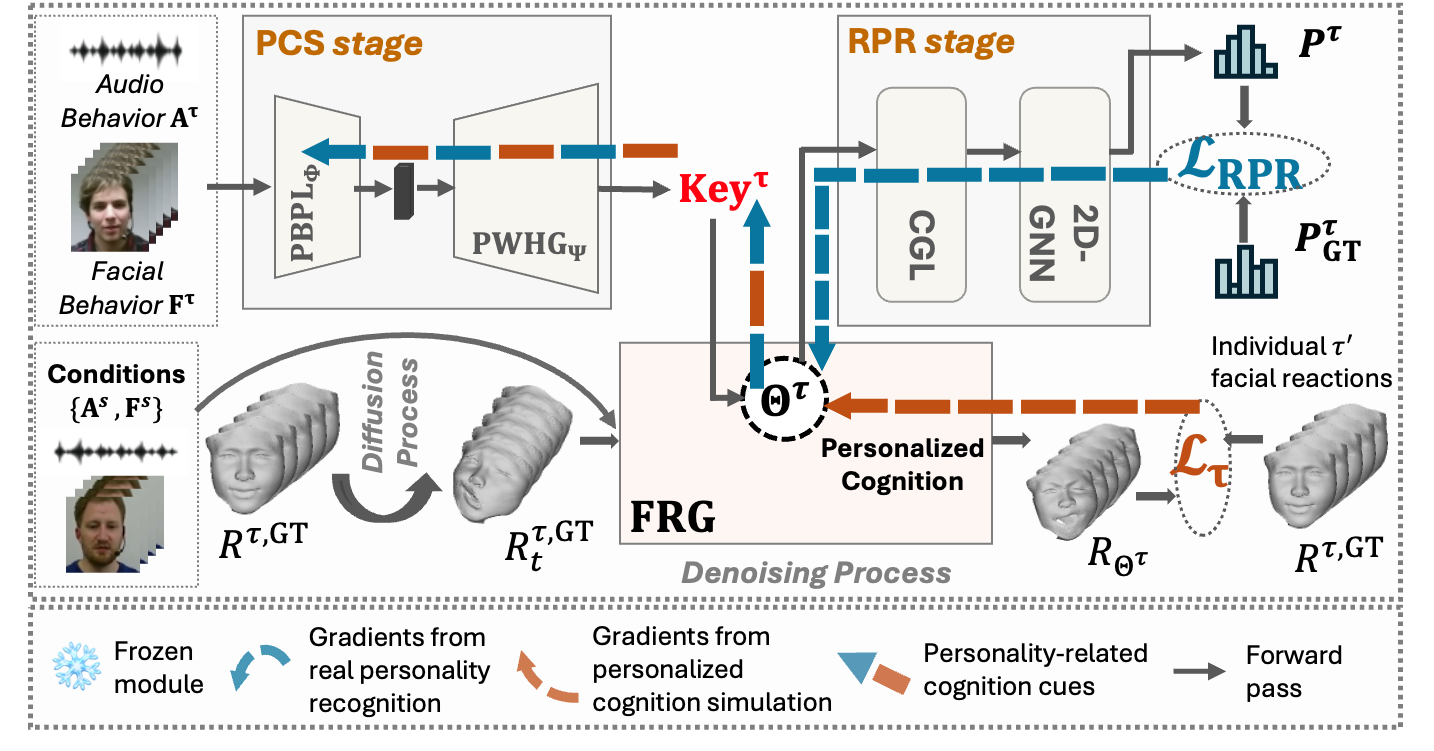}
  \vskip -0.1in
  \caption{Our end-to-end joint training strategy is driven by two loss functions: (1) $\mathcal{L}_{\tau}$ compares the generated facial reaction $R_{\Theta_{\tau}}$ with the GT personalised real facial reaction $R^{\tau, \text{GT}}$. The gradients are backpropagated to optimise $\text{\PersonalWeightGenerator}_{\Psi}$ and $\text{\personalisedbehaviourPatternLearning}_{\Phi}$; (2) $\mathcal{L}_{\textbf{RPR}}$ compares the personality prediction $\mathcal{P}^{\tau}$ with the GT real personality traits $\mathcal{P}^{\tau}_{\text{GT}}$. The gradients are back-propagated to train \MatrixGraphNeuralNetwork, CGL, $\text{\PersonalWeightGenerator}_{\Psi}$ and $\text{\personalisedbehaviourPatternLearning}_{\Phi}$. This way, both losses jointly enforce our PCS to simulate personality-related cognition from input audio-visual behaviours.}
  \label{fig:training_strategy}
\end{figure}

\noindent We pre-train the pre-defined architecture \FacialReactionGenerator\ as a generic facial reaction generator $\text{\FacialReactionGenerator}_{\Theta}$ in a diffusion manner with the objective loss function defined as: 
\begin{equation} 
\begin{split}
    &\mathcal{L}_{\mathrm{diff}} = \mathbb{E}_{t \in [1,T], R_{t}^\text{GT} \sim q_{t}}\left[ \left\| R^\text{GT} - R_\Theta \right\|_2^2 \right] \\
    & R_\Theta = \text{\FacialReactionGenerator}_{\Theta} (R_{t}^\text{GT}, t, \{\mathrm{A}^{s}, \mathrm{F}^{s} \})
\end{split}
\label{eq:diffusion-loss}
\end{equation}
where $R_{t}^\text{GT}$ is obtained by adding random Gaussian noises to the ground-truth (GT) facial reaction $R^\text{GT}$ via a forward diffusion process $q_t$ \cite{ho2020denoising}; $t$ is the diffusion step controlling the noise level; and $\{\mathrm{A}^{s}, \mathrm{F}^{s} \}$ are the input audio and facial behaviours) that triggered the corresponding real facial reaction $R^\text{GT}$. Since $R^\text{GT}$ is expressed by individuals with different personalities, $\text{\FacialReactionGenerator}_{\Theta}$ learns to simulate generic human cognition for generating various appropriate facial reactions that can be expressed by individuals of different personalities. 

Based on the pre-trained $\text{\FacialReactionGenerator}_{\Theta}$, our end-to-end joint training strategy is achieved by two loss functions. As illustrated in Fig. \ref{fig:training_strategy}, \textbf{the first loss $\mathcal{L}_{\tau}$} compares the facial reaction $R_{\hat{{\Theta^\tau}}}$ generated by the refined personalised facial reaction generator $\text{\FacialReactionGenerator}_{\Theta^\tau}$ with the corresponding real personalised appropriate facial reaction $R^{\tau, \text{GT}}$ expressed by $\tau$ (i.e., any facial reaction expressed by $\tau$ that is appropriate in response to the input $\mathrm{A}^{s}, \mathrm{F}^{s}$) \cite{zhu2024perfrdiff} as: 
\begin{equation}
\begin{split}
    & \mathcal{L}_{\tau} = \mathbb{E}_{t \in [1, T], R^{\tau, \text{GT}}_t \sim q_{t}} \left[ \left\| R^{\tau, \text{GT}} - R_{\Theta^{\tau}}  \right\|_2^2  \right] \\
    & R_{\Theta^\tau} = \text{\FacialReactionGenerator}_{\Theta^\tau} (R^{\tau, \text{GT}}_t, t, \{\mathrm{A}^{s}, \mathrm{F}^{s} \})
\end{split}
\end{equation}
where $R^{\tau, \text{GT}}_t$ is obtained by adding random Gaussian noises to $R^{\tau, \text{GT}}$ via a forward diffusion process $q_t$. During the training, the $\text{FRG}_{\Theta}$ are kept frozen, while the $\text{\personalisedbehaviourPatternLearning}_{\Phi}$ and $\text{\PersonalWeightGenerator}_{\Psi}$ modules generate the personalised key $\text{Key}^{\tau}$ to refine (i.e., via a differentiable adding operation) the weights 
of $\text{\FacialReactionGenerator}_{\Theta}$ as $\text{\FacialReactionGenerator}_{\Theta^{\tau}}$ that aims at recovering personalised facial reaction $R^{\tau, \text{GT}}$ from $R^{\tau, \text{GT}}_t$. This way, the gradients computed from the loss between the generated personalised facial reaction and the ground-truth $R^{\tau, \text{GT}}$ are back-propagated through the frozen $\text{\FacialReactionGenerator}_{\Theta}$ and $\text{Key}_{\tau}$ to optimise $\text{\personalisedbehaviourPatternLearning}_{\Phi}$ and $\text{\PersonalWeightGenerator}_{\Psi}$. Here, the gradient of $\mathcal{L}_{\tau}$ with respect to $\text{Key}_{\tau}$ is:
\begin{align}
    \xi^{\tau}_{n, k} &= \sigma (\Theta^{\tau}_{n,k}\ \xi^{\tau}_{n, k-1} + b_{n,k}), \nonumber \\
    \frac{\partial \mathcal{L}_{\tau}}{\partial \text{Key}^{\tau}_{n,k}} &= \frac{\partial \mathcal{L}_{\tau}}{\partial \xi^{\tau}_{n, k}} \cdot \frac{\partial \xi^{\tau}_{n, k}}{\partial \Theta^{\tau}_{n,k}} \cdot \frac{\partial \Theta^{\tau}_{n,k}}{\partial \text{Key}^{\tau}_{n,k}}, \nonumber \\
    &= \frac{\partial \mathcal{L}_{\tau}}{\partial \xi^{\tau}_{n, k}} \cdot \sigma^{\prime} \cdot \xi^{\tau}_{n, k-1} \cdot f_{\text{REF}}^{\prime}(\Theta, \text{Key}^{\tau}_{n,k})
\end{align}
where we denote $\xi^{\tau}_{n,k}$ as the intermediate activation generated from the $k$-th layer of the $n$-th block during the forward propagation of $\text{\FacialReactionGenerator}_{\Theta^{\tau}}$ (i.e., comprising the weight parameter $\Theta^{\tau}_{n,k}$ and the bias parameter $b_{n,k}$) while $\sigma^{\prime}$ denoting the derivative of the activation function $\sigma$. Since $\text{\FacialReactionGenerator}_{\Theta^{\tau}}$ is produced via the refinement function (defined in Eq. \ref{eq:key-generation}) based on the personalised $\text{Key}^{\tau}_{n,k}$,  $f_{\text{REF}}^{\prime}(\Theta, \text{Key}^{\tau}_{n,k})$ represents its derivative with respect to $\text{Key}^{\tau}_{n,k}$.

Meanwhile, \textbf{the second loss $\mathcal{L}_{\textbf{RPR}}$} compares the personality traits $\mathcal{P}^{\tau}$ predicted by our \MatrixGraphNeuralNetwork\ and the ground-truth real personality traits $\mathcal{P}^{\tau}_{\text{GT}}$ as:
\begin{equation}
    \mathcal{L}_{\textbf{RPR}} = \| \mathcal{P}^{\tau} - \mathcal{P}^{\tau}_{\text{GT}} \|^2
\end{equation}
More training and loss function details are provided in Supplementary Material.


\begin{table*}[h!]
    \centering
    \resizebox{\linewidth}{!}{ 
    \begin{tabular}{ccccccc | ccccccc}
        \toprule
        \multicolumn{7}{c|}{\textbf{UDIVA - MSE ($\downarrow$)}} & \multicolumn{7}{c}{\textbf{UDIVA - CCC ($\uparrow$)}} \\
        \midrule
        \textbf{Methods} & \textbf{Open} & \textbf{Cons} & \textbf{Extrav} & \textbf{Agree} & \textbf{Neuro} & \textbf{Avg.} & \textbf{Methods} & \textbf{Open} & \textbf{Cons} & \textbf{Extrav} & \textbf{Agree} & \textbf{Neuro} & \textbf{Avg.} \\
        \midrule
        TPN \cite{Yang_2020_CVPR} & 0.9259 & 0.7596 & 1.2524 & 0.9458 & 1.4147 & 1.0597 & TPN \cite{Yang_2020_CVPR} & 0.0448 & 0.0438 & 0.0287 & -0.0177 & -0.0281 & 0.0125 \\
        DCC \cite{guccluturk2016deep} & 0.9637 & 0.6709 & 1.5770 & 1.0364 & 1.5126 & 1.0801 & DCC \cite{guccluturk2016deep} & 0.0648 & 0.2424 & 0.0349 & 0.0305 & 0.0009 & 0.0743 \\
        CAM-DAN+ \cite{ventura2017interpreting} & 1.0223 & \textbf{0.5471} & 1.4800 & 0.8446 & 1.1493 & 1.0087 & CAM-DAN+ \cite{ventura2017interpreting} & 0.0336 & \textbf{0.3359} & 0.0270 & 0.2014 & 0.2709 & 0.1738 \\
        DAN \cite{8066355} & 1.0841 & 1.1174 & 2.0979 & 1.4846 & 1.2400 & 1.4048 & DAN \cite{8066355} & 0.0009 & 0.0087 & 0.0061 & 0.0035 & -0.0016 & 0.0035 \\
        DAN-MFCC \cite{8066355} & 1.1657 & 0.7358 & 2.0523 & 1.1006 & 1.0745 & 1.2258 & DAN-MFCC \cite{8066355} & 0.0008 & 0.1154 & 0.0132 & -0.0022 & 0.1219 & 0.0498 \\
        PersEmoN \cite{8897617} & 0.9034 & 0.6985 & 1.4097 & 0.9446 & 1.1469 & 1.0206 & PersEmoN \cite{8897617} & -0.0095 & 0.0133 & 0.0045 & 0.0058 & 0.0045 & 0.0037 \\
        CR-Net \cite{li2020cr} & 1.0985 & 0.6877 & 1.3079 & 0.9099 & 1.4140 & 1.0834 & CR-Net \cite{li2020cr} & 0.0998 & 0.1788 & 0.1158 & 0.2168 & 0.0449 & 0.1312 \\
        Amb-Fac \cite{SUMAN2022107715} & 0.8529 & 0.6394 & 1.4212 & 0.9334 & 1.1410 & 0.9976 & Amb-Fac \cite{SUMAN2022107715} & 0.0532 & 0.1941 & 0.0442 & 0.0453 & 0.0204 & 0.0714 \\
        OCEAN-AI \cite{ryumina2024ocean} & 0.9391 & 0.6675 & 1.4956 & 0.9588 & 1.1590 & 1.0440 & OCEAN-AI \cite{ryumina2024ocean} & -0.0348 & 0.0468 & 0.0302 & 0.0397 & 0.0041 & 0.0171 \\
        \midrule
        \textbf{Ours} & \textbf{0.7492} & 1.1114 & \textbf{0.7166} & \textbf{0.7829} & \textbf{0.7026} & \textbf{0.8125} & \textbf{Ours} & \textbf{0.2257} & -0.0345 & \textbf{0.2208} & \textbf{0.2988} & \textbf{0.4431} & \textbf{0.2308} \\
        \midrule
        \multicolumn{7}{c|}{\textbf{NoXI - MSE ($\downarrow$)}} & \multicolumn{7}{c}{\textbf{NoXI - PCC ($\uparrow$)}} \\
        \midrule
        \textbf{Methods} & \textbf{Open} & \textbf{Cons} & \textbf{Extrav} & \textbf{Agree} & \textbf{Neuro} & \textbf{Avg.} & \textbf{Methods} & \textbf{Open} & \textbf{Cons} & \textbf{Extrav} & \textbf{Agree} & \textbf{Neuro} & \textbf{Avg.} \\
        \midrule
        DCC \cite{guccluturk2016deep} & 0.112 & 0.755 & 0.640 & 0.082 & 0.102 & 0.338 & Spectral \cite{8976305} & 0.135 & 0.246  & 0.265 & 0.192  & 0.277 & 0.223 \\
        NJU-LAMDA \cite{8066355} & 0.099 & 0.135 & 0.069 & 0.075 & 0.210 & 0.118 & CR-Net \cite{li2020cr} & 0.181 & 0.271  & 0.301 & 0.177 & 0.325 & 0.251 \\
        VAT \cite{Girdhar_2019_CVPR} & 0.186 & 0.081 & 0.088 & 0.068 & 0.152 & 0.115 & VAT \cite{Girdhar_2019_CVPR} & -0.177 & 0.484 & 0.131 & 0.448 & 0.187 & 0.215 \\
        Amb-Fac \cite{SUMAN2022107715} & 0.094 & 0.047 & 0.035 & 0.038 & 0.049 & 0.053 & Amb-Fac \cite{SUMAN2022107715} &  -0.127 & 0.109 & 0.059 & 0.107 & 0.050 & 0.040 \\
        OCEAN-AI \cite{ryumina2024ocean} & 0.081 & 0.046 & 0.040 & 0.043 & \textbf{0.044} & 0.051 & OCEAN-AI \cite{ryumina2024ocean} & -0.039 & 0.305 & 0.281 & 0.331 & 0.201 & 0.216 \\
        Dyadformer \cite{Curto_2021_ICCV} & \textbf{0.063} & 0.049 & 0.044 & 0.048 & 0.050 & 0.051 & Dyadformer \cite{Curto_2021_ICCV} & 0.128 & 0.317 & 0.049 & 0.373 & 0.253 & 0.224 \\
        P-NAS \cite{9993801} & - & - & - & - & - & - & P-NAS \cite{9993801} & \textbf{0.189} & 0.376 & 0.420 & 0.289 & \textbf{0.481} & \textbf{0.351} \\
        \midrule
        \textbf{Ours} & 0.074 & \textbf{0.052} & \textbf{0.024} & \textbf{0.037} & 0.053 & \textbf{0.048} & \textbf{Ours} & 0.099 & \textbf{0.520} & \textbf{0.424} & \textbf{0.515} & 0.061 & 0.324 \\
        \bottomrule
    \end{tabular}
    }
    \caption{Comparison between our approach with most recent audio-visual personality recognition approaches on UDIVA and NoXI datasets, where UDIVA experiments are conducted following the widely-used benchmark \cite{10428080}, and thus \cite{Curto_2021_ICCV} is not compared as it specifically considered verbal/topic contexts.}
    \label{tab:exp-comparison}
\end{table*}

\section{Experiments}
\label{sec:exper}

\subsection{Experimental settings}
\textbf{Datasets}: We evaluate our method on two publicly available audio-visual datasets rigorously labeled with clip-level self-reported Big-Five real personality traits (Extraversion (Ex), Agreeableness (Ag), Openness (Op), Conscientiousness (Co), and Neuroticism (Ne)), including: (1) NoXI \cite{cafaro17_icmi} is a multilingual dyadic interaction dataset designed to study adaptive behaviours in “Expert-Novice” roles through video conferencing, encompassing 84 sessions across various topics; and (2) UDIVA \cite{palmero2021context}, comprising 188 in-person dyadic interactions from 147 participants recorded under four different tasks. 

\noindent \textbf{Implementation details}: We apply the pre-trained wav2vec 2.0 \cite{baevski2020wav2vec} to extract frame-level (25 fps) audio behaviour features, and a 3D Morphable Face Model (3DMM) \cite{blanz2023morphable,wang2022faceverse} to extract frame-level facial coefficients. The number of transformer layers within our \FacialReactionGenerator\ is empirically set to $N=9$. The default length of the audio-facial behaviour clips fed to \personalisedbehaviourPatternLearning\ is $10$ seconds. More dataset and implementation details are provided in Supplementary Material.

\noindent \textbf{Evaluation Metrics}: Following \cite{10428080}, we employ Mean Squared Error (MSE), Pearson Correlation Coefficient (PCC) and Concordance Correlation Coefficient (CCC) to evaluate personality recognition performances. 
As previous studies did not report PCC on UDIVA or CCC on NoXI, we present our results for these metrics in Supplementary Material.

\subsection{Comparison with existing approaches} 

\begin{figure*}[h]
  \centering
  \includegraphics[width=0.95\linewidth]{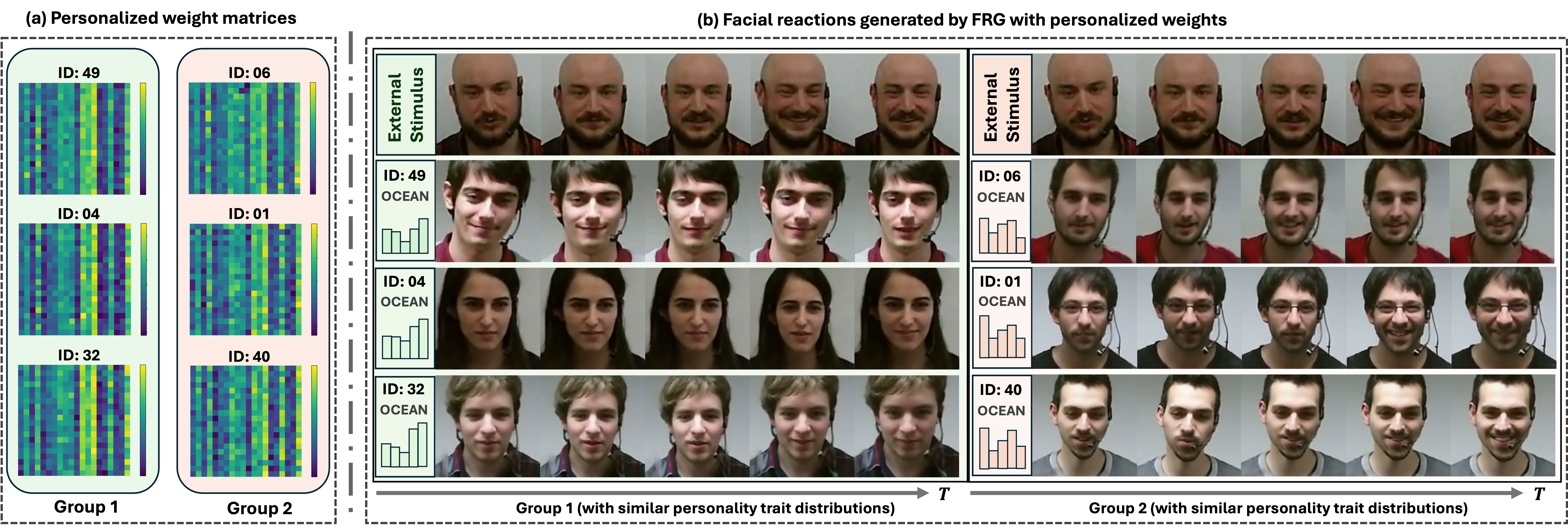}
  \caption{(a) Example personalised weight matrices learned for different individuals; (b) Individuals' facial reactions generated by \FacialReactionGenerator\ with different personalised weights in response to the same stimulus.}
  \label{fig:visualization}
\end{figure*}

Table \ref{tab:exp-comparison} shows that our cognition simulation approach clearly outperformed existing standard personalty recognition methods on UDVIA datasets, achieved 18.6\% and 32.8\% average MSE and CCC improvements over the previous state-of-the-arts (SOTA) \cite{SUMAN2022107715,ventura2017interpreting}. Our approach is particularly effective in inferring Extraversion and Agreeableness traits-related personalised cognition, as it achieved the best CCC/MSE results for them on both datasets. We explain this as these two traits may be more related to human external expressive behaviours \cite{lepri2012connecting,graziano2009agreeableness}. For the NoXI dataset, our approach achieved the best MSE results on 3/5 OCEAN traits (i.e., Conscientiousness, Extraversion and Agreeableness) as well as the averaged MSE (5.9\% relative improvements over the previous SOTA). It also achieved the second best CCC performance (29.1\% relative average CCC improvement over the third best method) with small disadvantages compared to the impractical method \cite{9993801} but \textit{over 2,000 times faster inference speed} (Table \ref{tab:time-comparison}). Importantly, our approach not only achieved clear superior personality recognition results but also experienced similar computational costs compared to existing standard personality recognition approaches. Fig. \ref{fig:visualization} shows that individuals with similar personality trait distributions (grouped by three per set) exhibit similar weight patterns learned from their audio-facial behaviours. Additionally, these individuals display comparable facial reactions to the same external stimulus (e.g., the left group shows reserved facial reactions, while the right group is more animated), highlighting the ability of our approach to capture personality-related cognition.

\begin{table}[h]
    \centering
    \resizebox{0.98\linewidth}{!}{ 
    \begin{tabular}{c|c|c} 
        \toprule
        \textbf{Methods} & \textbf{Inference Time (per $\tau$)} & \textbf{NoXI-PCC (Avg.)} \\
        \midrule
        OCEAN-AI \cite{ryumina2024ocean} & 1.3 minutes & 0.216 \\
        DCC \cite{guccluturk2016deep} & 1.8 minutes & 0.008 \\
        Dyadformer \cite{Curto_2021_ICCV} & 4.46 minutes & 0.224 \\
        Amb-Fac \cite{SUMAN2022107715} & 3.20 minutes & 0.040 \\
        \hline
        P-NAS \cite{9993801} & 8.3 days & 0.351 \\
        \textbf{Ours} & 5.29 minutes & 0.324 \\
        \bottomrule
    \end{tabular} 
    }
    \caption{Inference time of RPR models on a 10-minute audio-visual clip, where P-NAS \cite{9993801} and ours are cognition-based methods, while others play the role of external observers.}
    \label{tab:time-comparison}
\end{table}

\subsection{Ablation studies}
This section conducts a series of ablation studies on NoXI dataset. Additionally experiments are provided in Supplementary Material, comparing: (1) input modality contributions; (2) various training strategies, and investigate: (3) parameter sensitivity for key model components; (4) long-term stability of learned personality; (5) facial reaction generation results; (6) statistical significance; (7) model computational costs; and (8) model's bias and fairness.

\begin{table}[h]
    \centering
    \begin{adjustbox}{width=\columnwidth,center}
        \begin{tabular}{l c c c c c c c}
            \toprule
            \textbf{Methods} & \textbf{Metric} & \textbf{Open} & \textbf{Cons} & \textbf{Extrav} & \textbf{Agree} & \textbf{Neuro} & \textbf{Avg.} \\
            \midrule
            \multirow{2}{*}{$\text{Key}^{\tau}$ (w/o $\Theta$)} & \textbf{MSE} & 0.075 & 0.067 & 0.034 & 0.064 & \textbf{0.046} & {0.057} \\ 
            & \textbf{PCC} & 0.005 & 0.240 & 0.258 & 0.241 & \textbf{0.246} & {0.198} \\ 
            \midrule
            \multirow{2}{*}{$\text{Key}^{\tau} \times \Theta$} & \textbf{MSE} & \textbf{0.058} & 0.055 & 0.034 & \textbf{0.037} & 0.053 & \textbf{0.048} \\
            & \textbf{PCC} & \textbf{0.248} & 0.447 & 0.170 & 0.349 & -0.100 & 0.223 \\
            \midrule
            \multirow{2}{*}{$\text{Key}^{\tau} + \Theta$} & \textbf{MSE} & 0.074 & \textbf{0.052} & \textbf{0.024} & \textbf{0.037} & 0.053 & {\textbf{0.048}} \\ 
            & \textbf{PCC} & 0.099 & \textbf{0.520} & \textbf{0.424} & \textbf{0.515} & 0.061 & {\textbf{0.324}} \\ 
            \bottomrule
        \end{tabular}
    \end{adjustbox}
    \caption{Results achieved for different personalised cognition refinement strategies, where $\text{Key}^{\tau}$ (w/o $\Theta$) denotes directly generating $\text{Key}^{\tau}$ as the personalised cognition.}
    \label{tab:ablation-indexing}
\end{table}

\begin{table}[!ht]
    \label{tab:ablation_matrix_graph}
    \centering
    \begin{adjustbox}{width=\columnwidth,center}
        \begin{tabular}{c c c c c c c c}
            \toprule
            \textbf{Setting} & \textbf{Metric} & \textbf{Open} & \textbf{Cons} & \textbf{Extrav} & \textbf{Agree} & \textbf{Neuro} & \textbf{Avg.} \\
            \midrule
            Layer-level node & \textbf{MSE} & \textbf{0.056} & 0.067 & 0.046 & 0.062 & \textbf{0.047} & 0.056 \\
            (fully connected) & \textbf{PCC} & 0.214 & 0.590 & 0.161 & 0.400 & 0.004 & 0.274 \\ 
            \midrule
            Layer-level node & \textbf{MSE} & 0.070 & 0.055 & 0.036 & 0.052 & 0.054 & 0.053 \\
            (head-to-tail) & \textbf{PCC} & 0.105 & 0.456 & 0.223 & 0.457 & \textbf{0.213} & 0.291 \\ 
            \midrule
            Block-level node  & \textbf{MSE} & 0.059 & 0.061 & 0.036 & 0.049 & 0.052 & 0.052 \\
            (KNN) & \textbf{PCC} & \textbf{0.320} & 0.586 & 0.138 & 0.422 & 0.118 & 0.317 \\ 
            \midrule
            Block-level node & \textbf{MSE} & 0.074 & 0.052 & \textbf{0.024} & \textbf{0.037} & 0.053 & {\textbf{0.048}} \\ 
            (fully connected) & \textbf{PCC} & 0.099 & 0.520 & \textbf{0.424} & \textbf{0.515} & 0.061 & {\textbf{0.324}} \\ 
            \bottomrule
        \end{tabular}
    \end{adjustbox}
    \caption{Results achieved for different node definition/edge connection strategies. \textbf{Layer-level node:} treating each weight matrix of linear layer in each transformer block as a node; \textbf{head-to-tail:} all nodes within each transformer block are fully connected while the node of the last weight matrix within a block connects to the node of the first weight matrix in the following block; and \textbf{KNN:} empirically connecting each node to its six nearest neighbours (measured by L2 distance).}
    \label{tab:ablation-connection}
\end{table}

\textbf{Influences of model components:} Table \ref{tab:ablation-main} shows that directly inferring real personality from external behaviours in the short audio-facial clip (V1) leads to very poor performance. In contrast, with the help of \FacialReactionGenerator\ to simulate personalised cognition, the performance largely improved (V2), where pre-training \FacialReactionGenerator\ plays a crucial role (V8 vs V2 / V9 vs V3), suggesting that the simulated cognition provides informative personality cues. Meanwhile, our transformer block-level node encoding strategy (V7 vs V5), cognition graph representation learning (node feature matrices (V5 vs V4) and edge feature matrices (V9 vs V7)), and the corresponding 2D-GNN additionally enhanced the performance. This validates that such a cognition-based graph managed to preserve crucial structural cues regarding personality-related cognition, where edge features facilitates better task-specific message exchanging among node features. Finally, our end-to-end training strategy consistently enhanced the performance, indicating its effectiveness in simulating personality-related cognition.

\begin{table}[t]
    \centering
    \begin{adjustbox}{width=\linewidth,center}
        \begin{tabular}{c c c | c | c c | c | c c} 
            \toprule
            & \multicolumn{2}{c|}{\textbf{PCS stage}} & \multicolumn{3}{c|}{\textbf{RPR stage}} & \textbf{Training} & \multirow{3}{*}{\textbf{Metric}} & \multirow{3}{*}{\textbf{Avg.}} \\
            \cmidrule(lr){2-7} 
            & \multirow{2}{*}{$\text{FRG}_{\Theta}$} & $\text{FRG}_{\Theta}$ & Block-level & Matrix & Matrix & End-to-end & & \\
            & & Pre-training & nodes & nodes & edges  & training & & \\ 
            \midrule
            \multirow{2}{*}{V1} & & & \multirow{2}{*}{$\checkmark$} & \multirow{2}{*}{$\checkmark$} & \multirow{2}{*}{$\checkmark$} & & \textbf{MSE} & 0.067 \\
            & & & & & & & \textbf{PCC} & 0.110 \\

            \cmidrule(lr){1-9}
            \multirow{2}{*}{V2} & \multirow{2}{*}{$\checkmark$} & & \multirow{2}{*}{$\checkmark$} & \multirow{2}{*}{$\checkmark$} & \multirow{2}{*}{$\checkmark$} & & \textbf{MSE} & 0.060 \\
            & & & & & & & \textbf{PCC} & 0.179 \\
            
            \cmidrule(lr){1-9}
            \multirow{2}{*}{V3} & \multirow{2}{*}{$\checkmark$} & & \multirow{2}{*}{$\checkmark$} & \multirow{2}{*}{$\checkmark$} & \multirow{2}{*}{$\checkmark$} & \multirow{2}{*}{$\checkmark$} & \textbf{MSE} & 0.057 \\
            & & & & & & & \textbf{PCC} & 0.198 \\

            \cmidrule(lr){1-9}
            \multirow{2}{*}{V4} & \multirow{2}{*}{$\checkmark$} & \multirow{2}{*}{$\checkmark$} & & & & \multirow{2}{*}{$\checkmark$} & \textbf{MSE} & 0.054 \\
            & & & & & & & \textbf{PCC} & 0.158 \\

            \cmidrule(lr){1-9}
            \multirow{2}{*}{V5} & \multirow{2}{*}{$\checkmark$} & \multirow{2}{*}{$\checkmark$} & & \multirow{2}{*}{$\checkmark$} & & \multirow{2}{*}{$\checkmark$} & \textbf{MSE} & 0.053 \\ 
            & & & & & & & \textbf{PCC} & 0.254 \\

            \cmidrule(lr){1-9}
            \multirow{2}{*}{V6} & \multirow{2}{*}{$\checkmark$} & \multirow{2}{*}{$\checkmark$} & \multirow{2}{*}{$\checkmark$} & \multirow{2}{*}{$\checkmark$} & & & \textbf{MSE} & 0.052 \\
            & & & & & & & \textbf{PCC} & 0.197 \\ 

            \cmidrule(lr){1-9}
            \multirow{2}{*}{V7} & \multirow{2}{*}{$\checkmark$} & \multirow{2}{*}{$\checkmark$} & \multirow{2}{*}{$\checkmark$} & \multirow{2}{*}{$\checkmark$} & & \multirow{2}{*}{$\checkmark$} & \textbf{MSE} & 0.051 \\ 
            & & & & & & & \textbf{PCC} & 0.262 \\
            
            \cmidrule(lr){1-9}
            \multirow{2}{*}{V8} & \multirow{2}{*}{$\checkmark$} & \multirow{2}{*}{$\checkmark$} & \multirow{2}{*}{$\checkmark$} & \multirow{2}{*}{$\checkmark$} & \multirow{2}{*}{$\checkmark$} & & \textbf{MSE} & 0.051 \\
            & & & & & & & \textbf{PCC} & 0.319 \\
            
            \cmidrule(lr){1-9}
            \multirow{2}{*}{V9} & \multirow{2}{*}{$\checkmark$} & \multirow{2}{*}{$\checkmark$} & \multirow{2}{*}{$\checkmark$} & \multirow{2}{*}{$\checkmark$} & \multirow{2}{*}{$\checkmark$} & \multirow{2}{*}{$\checkmark$} & \textbf{MSE} & \textbf{0.048} \\
            & & & & & & & \textbf{PCC} & \textbf{0.324} \\
            \bottomrule
        \end{tabular}
    \end{adjustbox}
    \caption{Contributions of model components. \textbf{V1:} denotes no cognition simulation involved in prediction; \textbf{V4:} applying standard vector-based graph with GCN for processing simulated cognition; \textbf{V4/V5:} encoding every weight matrix in every block to a node feature vector/matrix.}
    \label{tab:ablation-main}
\end{table}

\textbf{Results of different 'Refine' operations:} Table \ref{tab:ablation-indexing} compares different strategies to project the learned $\text{Key}^\tau$ as the personalised cognition $\Theta^\tau$. It is clear that directly projecting $\text{Key}^\tau$ as $\Theta^\tau$ without considering the pre-learned $\Theta$ resulted in poor performance, i.e., \textit{it is difficult to directly obtain reliable personalised cognition from a short clip} (also evidenced by poor facial reaction generation results in Supplementary Material). Besides, simply adding $\text{Key}^\tau$ to $\Theta$ achieved superior results over applying it to weight $\Theta$.

\textbf{Results of different cognition graph node/topology settings:} As compared in Table \ref{tab:ablation-connection}, when representing each linear layer's weight matrix within each transformer block as a node feature, the performances are not as good as our default block-level node feature setting, regardless of their topology. More importantly, our fully connected topology outperformed the topology defined by the widely-used KNN strategy, suggesting that real personality-related cognition patterns may be reflected by the combination of all simulated cognitive processes (transformer blocks/nodes).

\section{Conclusion}
This paper presents an end-to-end audio-facial personalised human cognition simulation strategy for RPR. Results show that our strategy can effectively and efficiently simulate personality-related personalised cognition while our novel cognition graph learning and 2D-GNN also positively contributed to RPR. Key limitations include the incomplete integration of nuanced verbal features and the simple network architecture, which will be addressed in our future work. Our novel framework can be potentially extended for understanding other human internal status (e.g., expressive behaviour-based mental health/emotion analysis).

\bibliography{main}

\end{document}